# Quantitative coronary calcification analysis for prediction of myocardial ischemia using non-contrast CT calcium scoring

**Juhwan Lee[1,2], Sadeer Al-Kindi[3], Ammar Hoori[2], Tao Hu[2], Hao Wu[2], Justin N. Kim[2], Robert Gilkeson[4], Sanjay Rajagopalan[4], David L. Wilson[2,5,*]**

[1] *Department of Biomedical Engineering, Virginia Commonwealth University, Richmond, VA, 23284, USA*
[2] *Department of Biomedical Engineering, Case Western Reserve University, Cleveland, OH, 44106, USA*
[3] *DeBakey Heart and Vascular Institute, Houston Methodist Hospital, Houston, TX, 77030*
[4] *Harrington Heart and Vascular Institute, University Hospitals Cleveland Medical Center, Cleveland, OH, 44106, USA*
[5] *Department of Radiology, Case Western Reserve University, Cleveland, OH, 44106, USA*

*Corresponding author: dlw@case.edu
Telephone number: 216-368-4099

## Abstract

Non-contrast computed tomography calcium scoring (CTCS) is widely recognized as an effective tool for cardiovascular risk stratification. This study aimed to develop a novel machine learning framework for predicting myocardial ischemia from routine non-contrast CTCS scans using quantitative coronary calcium assessment. This study analyzed 1,375 patients who underwent both non-contrast CTCS and regadenoson stress cardiac positron emission tomography myocardial perfusion imaging within one year at University Hospitals Cleveland Medical Center. A total of 74 variables, including clinical variables, Agatston score, and calcium-omics features, were evaluated. Relevant features were identified using XGBoost with Shapley Additive exPlanations (SHAP). Predictive models were trained and evaluated using 5-fold cross-validation. Among 987 patients, 89 (9%) were positive for myocardial ischemia. The final model incorporated the Agatston score, eight calcium-omics features, and age. The proposed model achieved a precision of 98.9±3.0%, sensitivity of 79.2±8.4, and F1 score of 87.7±5.3%. The addition of calcium-omics features significantly improved predictive performance compared with models using clinical variables alone or clinical variables with the Agatston score ($p$<0.05). Interestingly, the number of calcified arteries, despite being the lowest-ranked feature based on SHAP analysis, showed the strongest association with myocardial ischemia in logistic regression analysis (odds ratio: 3.63, 95% confidence interval: 2.80-4.77, $p$<0.00001). We developed a machine learning approach for predicting myocardial ischemia using routinely acquired non-contrast CTCS scans. Calcium-omics features provided incremental predictive value beyond conventional risk factors and Agatston scoring and may support more accessible cardiovascular risk stratification.

## Abbreviations

- CAD: Coronary Artery Disease
- PET: Positron Emission Tomography
- CTCS: Computed Tomography Calcium Scoring
- CAC: Coronary Artery Calcification
- HU: Hounsfield Units
- SHAP: SHapley Additive exPlanations
- AUROC: Area Under the Receiver Operating Characteristic Curve
- AUPRC: Area Under the Precision-Recall Curve
- OR: Odds Ratio
- CI: Confidence Interval
- SVM: Support Vector Machine
- RF: Random Forest

# 1 Introduction

Coronary artery disease (CAD) remains the leading cause of morbidity and mortality worldwide [1, 2]. Among its clinical manifestations, myocardial ischemia, characterized by reduced coronary blood flow and insufficient oxygen supply to the myocardium, is associated with increased risk of adverse cardiovascular events and poor clinical outcomes [3]. Early detection of myocardial ischemia is critical for risk stratification, guiding therapeutic interventions, and preventing progression to myocardial infarction and heart failure. Various noninvasive imaging techniques have been developed to assess myocardial ischemia, including stress echocardiography, single-photon emission computed tomography, cardiac magnetic resonance imaging, and positron emission tomography (PET). Among these modalities, regadenoson stress cardiac PET perfusion imaging using $^{13}NH_3$ tracer is a well-established and highly accurate method for detecting myocardial ischemia because it enables quantitative assessment of myocardial perfusion and coronary flow reserve. However, stress PET imaging has several limitations, including radiation exposure, high cost, limited availability, and the need for pharmacologic stress agents and specialized imaging infrastructure.

Non-contrast computed tomography calcium scoring (CTCS) provides direct evidence of coronary atherosclerosis through the detection of calcified lesions in the coronary arteries. CTCS is widely recognized by clinical guidelines as an effective tool for cardiovascular risk stratification because coronary artery calcification (CAC) strongly correlates with overall atherosclerotic plaque burden [4, 5]. However, despite its established role in risk assessment, the utility of CTCS for predicting functional myocardial ischemia remains underexplored. Existing CTCS assessments primarily rely on the Agatston score [6], which may not fully capture the complex spatial distribution, morphology, and heterogeneity of coronary calcifications associated with ischemic disease. To address these limitations, we previously developed quantitative imaging approaches based on calcium-omics [7] and fat-omics [8] analyses, demonstrating that detailed quantitative characterization of coronary calcifications and epicardial fat can predict findings on coronary CT angiography, including obstructive CAD classification [9, 10] and positive remodeling [11], as well as future adverse cardiovascular outcomes [7, 12, 13].

We hypothesize that quantitative measurements of coronary calcifications derived from CTCS are associated with myocardial ischemia. In this pilot study, we developed a novel machine learning framework to predict myocardial ischemia from low-cost or no-cost screening non-contrast CTCS scans using quantitative coronary calcification assessment (i.e., calcium-omics). We analyzed clinical variables, Agatston score, and calcium-omics features extracted from individual calcifications, arterial territories, and whole-heart regions to identify the most predictive imaging biomarkers of ischemia. To the best of our knowledge, this is the first study to investigate the prediction of myocardial ischemia directly from non-contrast CTCS imaging using quantitative coronary calcification analysis and machine learning.

# 2 Methods

### *2.1 Study cohort*

This pilot study retrospectively analyzed 1,375 patients who underwent both non-contrast CTCS and regadenoson stress cardiac PET myocardial perfusion imaging within one year between 2014 and 2020. The study cohort was derived from the CLARIFY registry at University Hospitals Cleveland Medical Center (ClinicalTrials.gov Identifier: NCT04075162) as part of a no-cost cardiovascular risk assessment program. This study was approved by the Institutional Review Board of University Hospitals Cleveland Medical Center with a waiver of informed consent for retrospective analysis. Inclusion criteria were: (1) availability of non-contrast CTCS scans, (2) reganenoson stress cardiac PET myocardial perfusion imaging performed within one year of CTCS acquisition, and (3) complete clinical and imaging data required for analysis. Exclusion criteria included: (1) prior coronary artery bypass graft surgery, (2) severe imaging artifacts or incomplete CTCS scans limiting quantitative calcium analysis, and (3) missing or nondiagnostic PET perfusion imaging results. In addition, 388 patients with zero CAC scores were excluded because of the low prevalence of myocardial ischemia in this subgroup (10 of 388, 2.6%). The final study population consisted of 987 patients. Figure 1 illustrates the overall workflow of the proposed framework for predicting myocardial ischemia using non-contrast CTCS scans.

*2.2 CTCS image acquisition*

Non-contrast CTCS scans were obtained using a Brilliance iCT scanner (Philips Healthcare) following a standardized clinical acquisition protocol (120 *kVp*, 30 *mAs*, 0.5 x 0.5 *mm* in-plane resolution, and 2.5 *mm* slice thickness). Images were reconstructed using soft-kernel filtered back projection. Coronary calcifications were identified and annotated by experienced cardiologists based on regions with attenuation values ≥130 Hounsfield Units (HU) within the coronary arteries, including the left main, left anterior descending, left circumflex, and right coronary arteries. For image preprocessing, voxel intensities were clipped to a range of -1024 t o1024 HU and normalized to a range of 0 to 1 to improve numerical stability for downstream analysis.

*2.3 Identification of myocardial ischemia using stress cardiac PET perfusion imaging*

Myocardial ischemia was assessed using regadenoson stress cardiac PET myocardial perfusion imaging with $^{13}NH_3$ tracer according to standard clinical protocols. Pharmacologic stress was induced using regadenoson, followed by intravenous administration of the $^{13}NH_3$ radiotracer for rest and stress myocardial perfusion imaging at rest and stress conditions. PET images were acquired using a dedicated PET/CT scanner with low-dose CT attenuation correction.

Myocardial perfusion images were reconstructed and interpreted by experienced nuclear cardiologists as part of routine clinical care. Myocardial ischemia was defined based on the presence of reversible perfusion defects on stress-rest PET perfusion imaging, consistent with inducible ischemia. Patients with abnormal stress perfusion findings indicative of ischemia were classified as ischemia-positive, whereas patients without evidence of inducible perfusion abnormalities were classified as ischemia-negative. Clinical PET imaging interpretations documented in the electronic health record were used as the reference standard for ischemia classification.

*2.4 Quantitative calcium-omics feature extraction*

To predict myocardial ischemia from CTCS scans, a comprehensive set of clinical, Agatston score, and calcium-omics features was extracted and analyzed. A total of 74 features were evaluated, including 4 clinical variables, the total Agatston score, and 69 calcium-omics features [7]. These features were designed to characterize calcified plaque burden and morphology across multiple anatomical and spatial scales associated with ischemic cardiovascular disease.

*Clinical variables* included baseline patient characteristics, including age, sex, diabetes mellitus, and smoking history. *The Agatston score*, a standard quantitative measure of CAC burden, was computed by multiplying the area of each calcified lesion by a density weighting factor based on peak attenuation values [6]. *Calcium-omics* [7] features consisted of quantitative imaging biomarkers extracted from individual calcified lesions, coronary artery territories, and whole-heart regions (Fig. 1). At the individual lesion level, extracted features included calcium mass, lesion volume, attenuation-based metrics derived from HU values (e.g., minimum, maximum, and mean HU), first- and second-order statistical descriptors, lesion shape characteristics, and spatial relationship measures such as distances between neighboring calcified lesions and relative anatomical location within the heart. At the coronary artery and whole-heart levels, aggregate quantitative descriptors were calculated to characterize the distribution and heterogeneity of calcified plaque burden, including mean, standard deviation, skewness, kurtosis, and histogram-derived attenuation statistics. Conventional CAC metrics, including Agatston score, calcium volume, and calcium mass, were additionally summarized at the lesion, vessel, and whole-heart levels to capture both localized and global patterns of coronary calcification. Detailed methods for calcium-omics feature extraction and validation have been previously reported [7].

### *2.5 Feature selection*

Relevant features were identified using the Extreme Gradient Boosting (XGBoost) algorithm [14] combined with SHapley Additive exPlanations (SHAP) method [15]. XGBoost was selected because of its strong performance in handling high-dimensional features and its ability to model nonlinear relationships and feature interactions.

The XGBoost classifier was configured using a binary logistic objective function and evaluated using the area under the receiver operating characteristics curve (AUROC). Model hyperparameters included a learning rate of 0.05, maximum tree depth of 3, minimum child weight of 3, subsample ratio of 0.6, and column subsampling ratio of 0.75. To reduce model overfitting and improve generalizability, both L1 (alpha = 0.5) and L2 (lambda = 5) regularization were applied, together with a minimum split loss reduction parameter (gamma = 0.5). The maximum number of boosting iterations was set to 1,000, and early stopping was implemented with a patience of 30 rounds based on validation AUC performance.

To enhance model robustness and minimize overfitting, a 5-fold cross-validation strategy was employed. The dataset was randomly divided into five mutually exclusive subsets, where, during each iteration, four subsets were used for model training, and the remaining subset was reserved for validation. All feature selection procedures were performed exclusively within the training folds to avoid data leakage.

Feature importance was subsequently quantified using the SHAP framework, which estimates the contribution of each feature to model predictions. For each fold, absolute SHAP values were computed for all features, and the mean absolute SHAP values were averaged across cross-validation folds to obtain a stable and interpretable ranking of feature importance.

### *2.6 Machine learning model development*

Predictive models were developed using the XGBoost algorithm [14], utilizing the most relevant features identified through the XGBoost-SHAP feature selection framework described in Section 2.5. XGBoost was selected because of its strong predictive performance, robustness to heterogenous clinical datasets, and ability to capture nonlinear relationships and higher-order feature interactions.

For model training, gradient-boosted decision trees were sequentially constructed to optimize predictive discrimination between positive and negative outcomes. The same XGBoost hyperparameters were consistently applied during both feature selection and model development to maintain methodological consistency and minimize variability between the selection and classification stages. Specifically, model optimization utilized a learning rate of 0.05 and a maximum tree depth of 3. To improve stability and reduce overfitting, row subsampling (0.6) and feature subsampling (0.75) were incorporated during tree construction. Additional regularization constraints included L1 regularization (alpha = 0.5), L2 regularization (lambda = 5), a minimum child weight of 3, and a minimum loss reduction parameter (gamma = 0.5). Model training was performed for up to 1,000 boosting iterations, while early stopping was applied if the validation AUC failed to improve for 30 consecutive rounds.

To investigate the incremental predictive contribution of quantitative calcium-omics features, three separate machine learning models were constructed and evaluated using different feature combinations. Model 1 incorporated clinical variables only. Model 2 combined clinical variables with the Agatston score. Model 3 integrated clinical variables, Agatston score, and calcium-omics features.

All models were evaluated using an identical 5-fold cross-validation framework to ensure a fair comparison among feature combinations. During each cross-validation iteration, the dataset was partitioned into training and validation subsets, and model training procedures were performed exclusively within the training folds to avoid data leakage. Model performance was assessed by averaging the validation metrics across all folds.

### *2.7 Performance evaluation*

Model performance was evaluated using a repeated 5-fold cross-validation framework to assess predictive robustness, stability, and generalizability. During each iteration, the dataset was randomly divided into five mutually exclusive subsets, where four subsets were used for model training and the remaining subset was reserved for validation. This process was repeated across all folds, and the resulting performance metrics were averaged to obtain overall model performance estimates while minimizing potential overfitting and sampling bias.

The predictive performance of each model was quantitatively assessed using multiple classification metrics, including precision, sensitivity, specificity, accuracy, F1 score, AUROC, and area under the precision-recall curve (AUPRC). For each metric, the mean and standard deviation across the cross-validation folds were calculated to evaluate both model performance and stability.

To investigate the association between individual variables and the study outcome, univariable logistic regression analyses were initially performed. Variables demonstrating statistical significance in the univariable analyses were subsequently incorporated into multivariable logistic regression models to identify independent predictors while accounting for potential confounding effects. Odds ratios (ORs), 95% confidence intervals (CIs), and corresponding p-values were reported for all regression analyses.

To further assess the robustness of the proposed XGBoost framework, its predictive performance was compared with several widely used machine learning algorithms, including Random Forest (RF), Support Vector Machine (SVM), and CatBoost. All comparison models were trained and evaluated using identical cross-validation partitions and evaluation procedures to ensure fair and unbiased performance comparisons.

*2.8 Statistical analysis*

Baseline clinical and imaging characteristics were summarized using appropriate descriptive statistics. Continuous variables were expressed as mean ± standard deviation for normally distributed data or median with interquartile range for non-normally distributed data, whereas categorical variables were presented as counts and percentages. Data normality was assessed using the Shapiro-Wilk test.

Comparisons between groups were performed using the independent Student's t-test for normally distributed continuous variables and the Mann-Whitney U test for non-normally distributed variables. Categorical variables were compared using the Chi-square test or Fisher's exact test when appropriate. A two-sided $p$-value <0.05 was considered statistically significant.

To compare predictive performance among machine learning models, the DeLong test was used for comparisons of AUROC and AUPRC values. McNemar's test was additionally applied for paired comparisons of classification results between competing models when appropriate. All statistical analyses were conducted using R software (version 2024.04.1; R Foundation for Statistical Computing, Vienna, Austria).

# 3 Results

Of the 987 patients included in this study, 89 patients were positive for myocardial ischemia. Baseline clinical characteristics are summarized in Table 1. There was no significant difference in age between the myocardial ischemia and non-myocardial ischemia groups (63.9±8.9 vs. 62.7±8.8 years). The proportion of female patients was significantly lower in the myocardial ischemia group compared with the non-myocardial ischemia group (30.3% vs. 53.7%, $p$=0.00003). Smoking status was not significantly different between groups (42.7% vs. 39.4%). However, baseline diabetes mellitus was more prevalent in patients with myocardial ischemia than in those without myocardial ischemia (40.4% vs. 23.5%, $p$=0.0008).

XGBoost-SHAP analysis was used to determine the most relevant variables for each predictive model. For Model 1, all four baseline clinical variables, including gender, diabetes mellitus, age, and smoking status, showed meaningful contributions in SHAP analysis (Fig. 2A) and were therefore included in training. In Model 2, after addition of the Agatston score, smoking status demonstrated minimal contribution and was excluded (Fig. 2B). As a result, Model 2 consisted of four variables: Agatston score, diabetes mellitus, gender, and age. For model 3, the top ten variables were selected empirically according to SHAP rankings (Fig. 2C). The Agatston score showed the strongest overall contribution, while eight of the ten selected features were derived from calcium-omics analysis, with age retained as the only clinical variable.

Classification performance improved substantially with the addition of imaging-derived features. Model 1, which included clinical variables only, showed limited predictive performance with sensitivity of 57.5±13.1% and overall accuracy of 62.2±4.7% (Table 2). The model achieved an AUROC of 0.620 and AUPRC of 0.166 (Fig 3). Incorporation of the Agatston score in Model 2 markedly improved performance, increasing the accuracy and F1 score to 95.7±1.3% and 76.4±6.7%, respectively, while achieving an AUROC of 0.912 and AUPRC of 0.822. Model 3, which additionally incorporated calcium-omics features, showed the highest performance with 98.0±1.0% accuracy, 87.7±5.3% F1 score, an AUROC of 0.931, AUPRC of 0.864 (Table 2 and Fig. 3).

In univariable logistic regression analysis, among the 10 selected features, multiple calcium-related features were significantly associated with myocardial ischemia, whereas age and massHist2 (histogram bin 2 of 5 lesion-based mass score bins) were not significantly associated with ischemia (Table 3). Notably, the number of calcified arteries (numArtCalc) showed the highest OR (3.63, 95% CI: 2.804-4.773), despite being the lowest-ranked feature based on SHAP analysis (Fig. 2C). In multivariable analysis, none of the selected features remained independently associated with myocardial ischemia. Agatston score demonstrated a neutral association after adjustment (adjusted OR: 1.001, 95% CI: 0.980-1.023, $p$=0.90), whereas numArtCalc showed the highest adjusted OR (1.32, 95% CI: 0.734-2.361) among the selected features; however, it did not reach statistical significance (Table 3).

Among the evaluated machine learning models, the XGBoost model consistently demonstrated the best overall classification performance for predicting myocardial ischemia, achieving the highest precision (98.9±3.0%), sensitivity (79.2±8.4), accuracy (98.0±1.0%), F1 score (87.7±5.3%) (Fig. 4). The RF model also showed strong performance with comparable accuracy and specificity, whereas the CatBoost model demonstrated the lowest overall performance. Nevertheless, all models exhibited relatively consistent classification performance with high specificity (>97%) and comparable sensitivity (78-79%). These findings suggest that the selected calcium-omics features are robust and reproducible across different machine learning algorithms, supporting their potential generalizability for myocardial ischemia prediction across diverse modeling approaches and datasets.

## 4 Discussions

Building on our extensive CT imaging studies [7, 9, 11, 12, 16–27], we developed a novel machine learning framework to predict myocardial ischemia from no- (or low-) cost screening non-contrast CTCS scans using quantitative coronary calcification assessment (i.e., calcium-omics). To our knowledge, this is the first study to investigate the feasibility of predicting myocardial ischemia from non-contrast CTCS imaging. Our study provides several important contributions. First, we demonstrated the incremental value of the Agatston score and, more importantly, calcium-omics features for myocardial ischemia prediction beyond conventional clinical risk factors. Second, calcium-omics features substantially improved model performance, achieving excellent discrimination and reproducible classification performance across multiple machine learning algorithms. Third, our findings suggest that routinely acquired CTCS scans may provide functional information related to myocardial ischemia in addition to conventional calcium burden assessment, potentially enabling more accessible and cost-effective cardiovascular risk stratification without additional imaging or radiation exposure.

Using XGBoost and SHAP analyses, our final model (Model 3) included the top 10 features, consisting of the Agatston score, eight calcium-omics features, and age. Among these variables, the Agatston score showed the strongest contribution to myocardial ischemia prediction, consistent with its established role as a marker of coronary atherosclerotic burden. In addition, calcium-omics features such as Area2D (total heart summation of lesion areas across all slices), mass score, massHist2 (histogram bin 2 of 5 lesion-based mass score bins), and volume score further improved classification performance beyond conventional clinical variables and Agatston scoring alone. Interestingly, the number of calcified arteries (numArtCalc) showed the lowest SHAP contribution but the highest adjusted OR in multivariable logistic regression. This discrepancy may reflect the distinct characteristics of machine learning-based feature importance and regression-based association analyses. SHAP values evaluate the contribution of a feature to overall model prediction within a nonlinear framework, whereas logistic regression assesses the independent effect of each variable after accounting for other covariates. Because many calcium-omics features are inherently intercorrelated, multicollinearity may have reduced the stability and statistical significance of regression coefficients. In addition, the relatively limited number of ischemic cases and potential feature-selection bias may also have contributed to variability between SHAP rankings and logistic regression findings.

Our findings may have important clinical implications for cardiovascular risk assessment and screening. Although the Agatston score is widely used to estimate coronary atherosclerotic burdens and future cardiovascular risk, our results suggest that calcium-omics features derived from routine non-contrast CTCS scans provide complementary information associated with myocardial ischemia. Because the proposed approach uses standard CTCS imaging, no additional contrast administration, imaging study, or radiation exposure is required. This suggests that non-contrast CTCS scans may provide not only structural information related to coronary calcium burden, but also indirect functional information associated with myocardial ischemia. If validated in larger multicenter studies, calcium-omics-based machine learning models may help identify patients who would benefit from additional functional testing or aggressive preventive management. Furthermore, automated calcium-omics analysis could be incorporated into existing CTCS workflows, potentially improving the clinical value of coronary calcium screening in routine practice.

This study has several limitations. First, this was a single-center retrospective study with a relatively limited sample size, including 987 patients and 89 ischemic cases. Although the proposed models demonstrated strong classification performance, the relatively small number of positive cases may have affected model stability and increased the risk of overfitting. Therefore, external validation in larger multicenter cohorts is necessary to confirm the generalizability and robustness of the proposed approach. Second, patients with zero coronary artery calcium were excluded from the analysis because the prevalence of myocardial ischemia in this subgroup was very low (10 of 388 patients, 2.6%). As a result, the current findings are limited to patients with detectable coronary calcification and may not be directly applicable to patients with zero CAC. Future studies with larger cohorts including more ischemic cases in the zero-CAC population are warranted.

In this study, we developed a novel machine learning approach for predicting myocardial ischemia using routinely acquired non-contrast CTCS scans and quantitative coronary calcium-omics assessment. Our findings showed that calcium-omics features provide incremental predictive value beyond traditional Agatston scoring and conventional clinical risk factors. If validated in larger multicenter studies, calcium-omics-based analysis may support more accessible and cost-effective cardiovascular risk stratification and help identify patients who may benefit from further functional evaluation.

## Acknowledgments

The content of this report is solely the responsibility of the authors and does not necessarily represent the official views of the National Institutes of Health. The grants were obtained via collaboration between Case Western Reserve University and University Hospitals of Cleveland. This work made use of the High-Performance Computing Resource in the Core Facility for Advanced Research Computing at Case Western Reserve University. The veracity guarantor, Justin N. Kim, affirms to the best of his knowledge that all aspects of this paper are accurate.

## Sources of Funding

This project was supported by the National Heart, Lung, and Blood Institute through grants K01HL171795, R01HL167199, and NIH R01HL165218. This research was conducted in space renovated using funds from an NIH construction grant (C06 RR12463) awarded to Case Western Reserve University.

## Tables

Table 1 Baseline clinical characteristics of patients with myocardial ischemia ($n$=89) and without myocardial ischemia ($n$=898).

| Clinical Features | Myocardial Ischemia ($n$=89) | Non-Myocardial Ischemia ($n$=898) | $p$-value |
|---|---|---|---|
| Age | 63.9±8.9 | 62.7±8.8 | 0.2648 |
| Female | 27 (30.3%) | 482 (53.7%) | 0.0000 |
| Smoking Status | 38 (42.7%) | 354 (39.4%) | 0.5711 |
| Diabetes Mellitus | 36 (40.4%) | 211 (23.5%) | 0.0008 |

Table 2 Classification performance for myocardial ischemia prediction using models trained with clinical variables only (Model 1), clinical variables with Agatston score (Model 2), and combined clinical, Agatston score, and calcium-omics features (Model 3).

| Models (XGBoost) | Precision | Sensitivity | Specificity | Accuracy | F1 |
|---|---|---|---|---|---|
| Model 1 (Clinical only) | 13.2±3.7 | 57.5±13.1 | 93.7±2.1 | 62.2±4.7 | 21.3±5.3 |
| Model 2 (Clinical + Ag) | 74.7±9.5 | 79.2±8.3 | 98.0±0.7 | 95.7±1.3 | 76.4±6.7 |
| Model 3 (Clinical + Ag + Calcium-omics) | 98.9±3.0 | 79.2±8.4 | 97.9±1.0 | 98.0±1.0 | 87.7±5.3 |

Table 3 Univariable and multivariable logistic regression analyses of the clinical, Agatston score, and calcium-omics features selected by XGBoost and SHAP for myocardial ischemia prediction.

| **Features** | **Univariable OR (95% CI)** | ***p*-value** | **Multivariable OR (95% CI)** | ***p*-value** |
|---|---|---|---|---|
| Age | 1.016 (0.991-1.043) | 0.213 | - | - |
| AgatstonScore2D | 1.012 (1.009-1.014) | $p<0.00001$ | 1.001 (0.981-1.023) | 0.904 |
| MassScore | 1.074 (1.058-1.089) | | 0.967 (0.885-1.059) | 0.461 |
| VolumeScore | 1.014 (1.011-1.017) | | 1.011 (0.977-1.033) | 0.393 |
| Area2D | 1.034 (1.027-1.043) | | 1.015 (0.946-1.104) | 0.697 |
| numArtCalc | 3.628 (2.804-4.773) | | 1.322 (0.734-2.361) | 0.347 |
| AgastonScorePerArtery2D_LM1 | 1.019 (1.013-1.025) | | 1.064 (0.956-1.211) | 0.372 |
| AgastonScorePerArtery3D_LM1 | 1.013 (1.007-1.019) | | 0.950 (0.843-1.026) | 0.423 |
| HUperArtery2D_stat_LCX3 | 1.009 (1.007-1.011) | | 1.000 (0.995-1.004) | 0.829 |
| massHist2 | 1.821 (0.812-3.812) | 0.127 | - | - |

**Figures with Figure Legends**

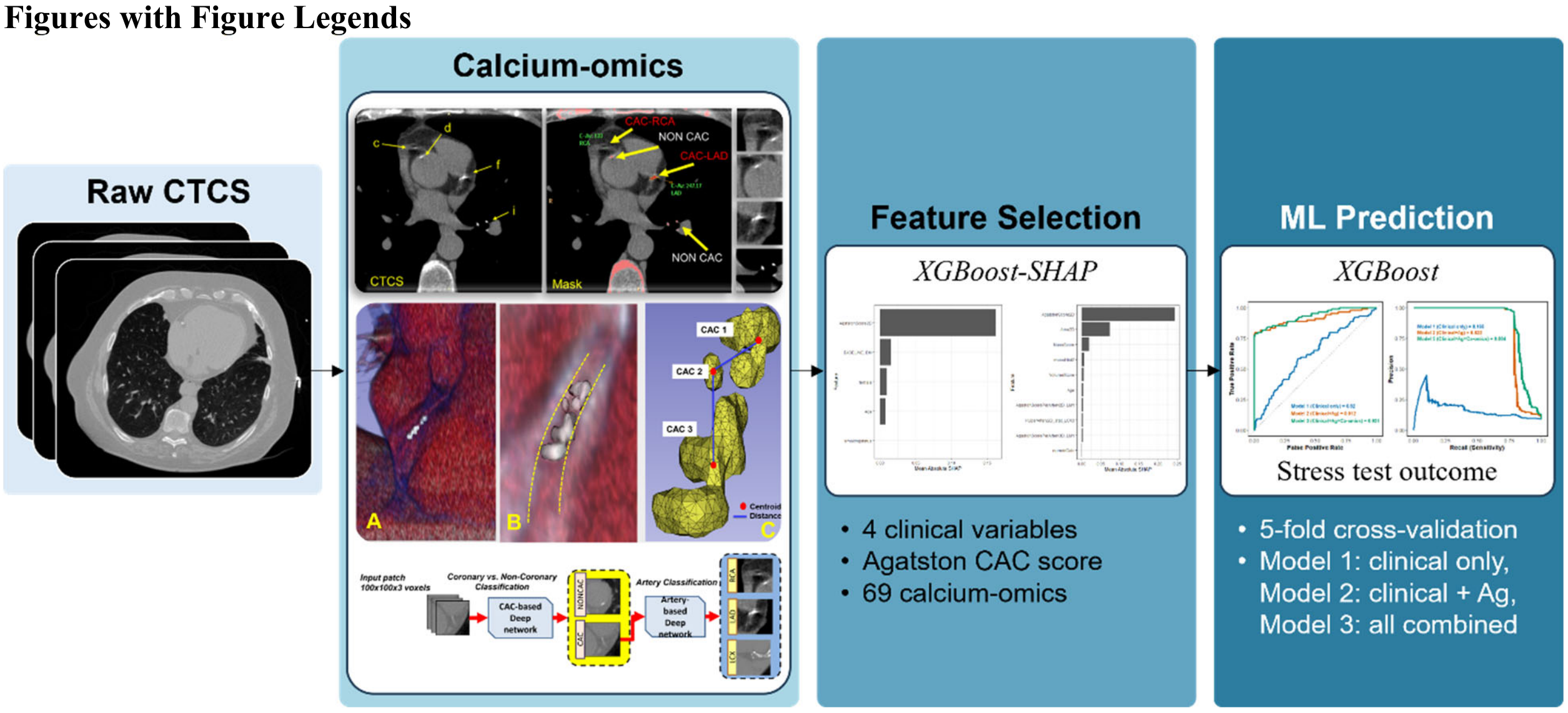

Fig. 1 Workflow of the proposed framework for predicting myocardial ischemia using routine non-contrast CTCS scans. The framework consisted of three major steps: feature extraction, feature selection, and machine learning-based prediction. A total of 75 features were extracted, including clinical variables, Agatston score, and calcium-omics features. Relevant features were identified using XGBoost with SHAP analysis and subsequently used to train predictive models for myocardial ischemia prediction. Model performance was evaluated using five-fold cross-validation with 1,000 repeated iterations.

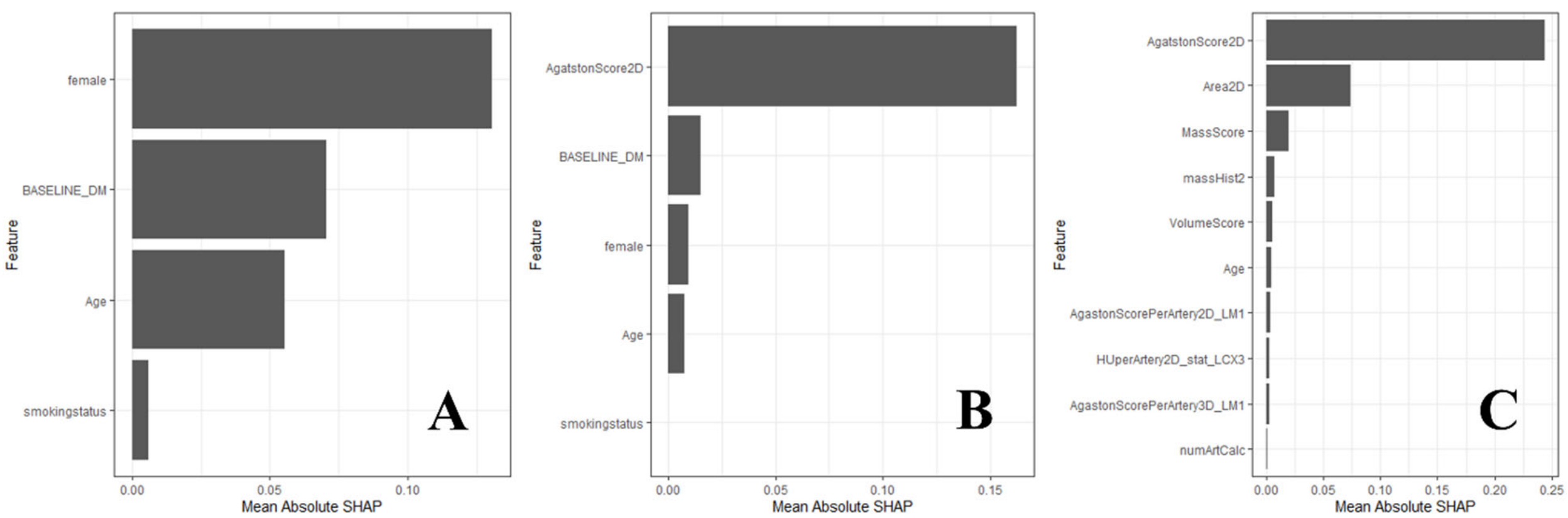

Fig. 2 Feature importance for each predictive model determined using XGBoost with SHAP analysis. Model 1 (A) included four clinical variables: gender, diabetes mellitus, age, and smoking status. In Model 2 (B), smoking status was excluded after incorporation of the Agatston score due to its minimal contribution. Model 3 (C) included the top 10 variables, consisting of the Agatston score, eight calcium-omics features, and age.

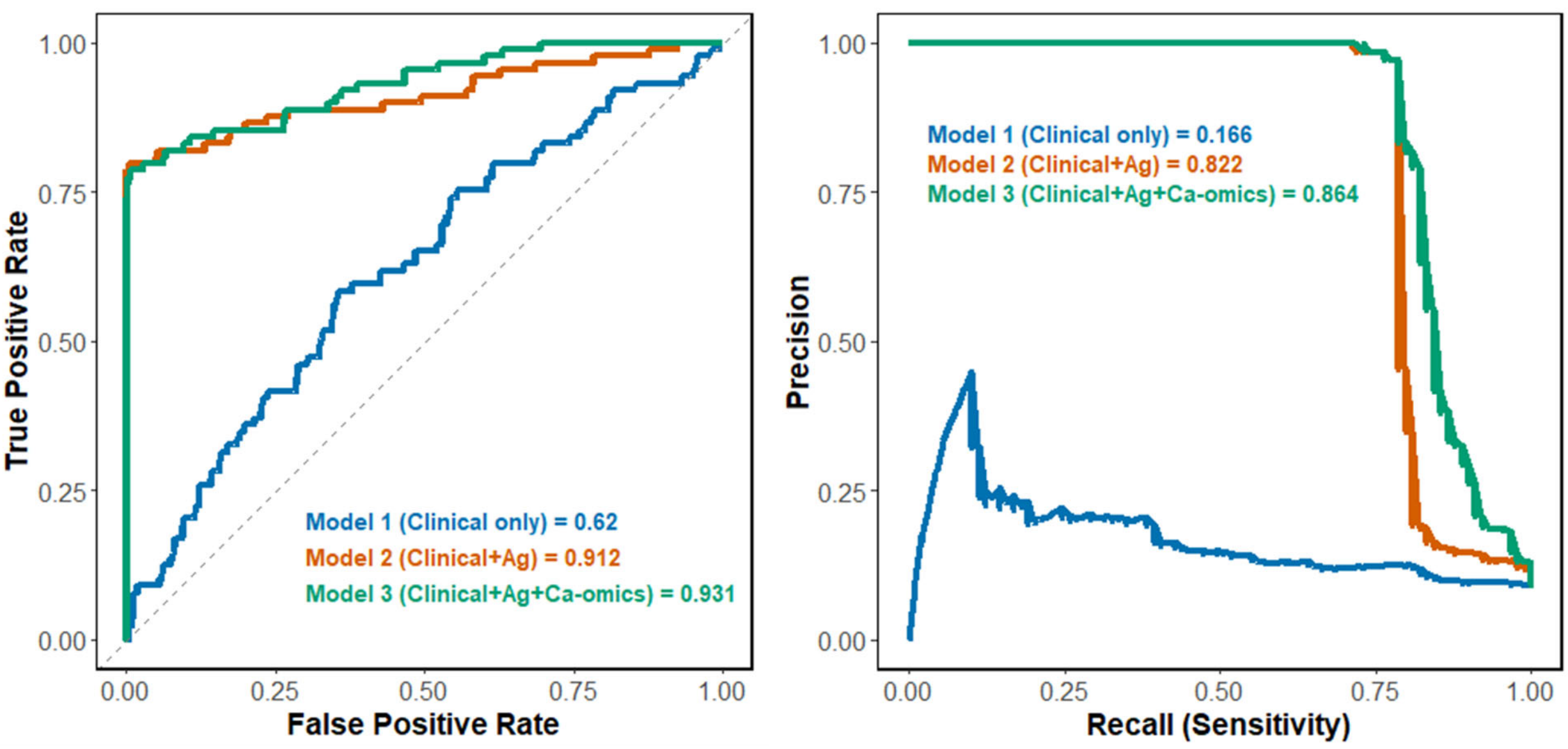

Fig. 3 AUROC (left) and AUPRC (right) for myocardial ischemia prediction using the three predictive models. Incorporation of the calcium-omics features substantially improved classification performance compared with the clinical variable with Agatston score model. The AUROC improved from 0.620 in Model 1 to 0.912 in Model 2 and 0.931 in Model 3, while the AUPRC increased from 0.166 to 0.822 and 0.864, respectively, demonstrating the incremental predictive value of quantitative coronary calcium assessment for myocardial ischemia prediction.

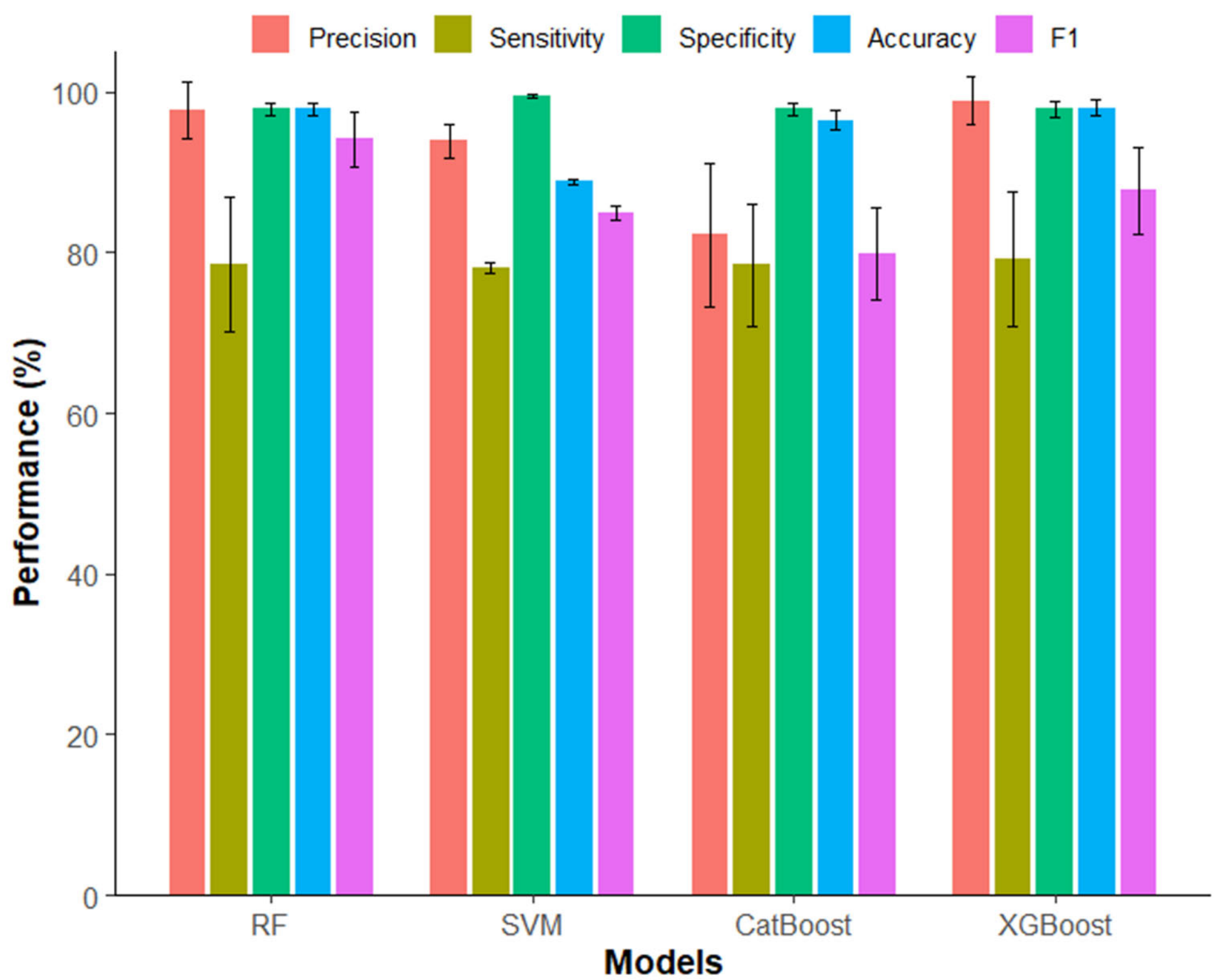

Fig 4 Comparison of myocardial ischemia classification performance among RF, SVM, CatBoost, and XGBoost models using the selected clinical variable, Agatston score, and calcium-omics features. XGBoost demonstrated the best overall performance. All models showed consistently high specificity (97.9±0.8, 99.5±0.2, 97.9±0.8, and 97.9±1.0 for RF, SVM, CatBoost, and XGBoost, respectively) and comparable sensitivity (78.6±8.4, 78.1±0.6, 78.5±7.6, and 79.2±8.4, respectively).